\documentclass[letterpaper, 10 pt, conference]{ieeeconf}  

\IEEEoverridecommandlockouts                              
\usepackage[utf8]{inputenc}
\usepackage[T1]{fontenc}
\usepackage{amsmath}
\usepackage{amssymb}
\usepackage{graphicx}
\usepackage{rotating}
\usepackage{tabularx} 
\usepackage{booktabs} 
\usepackage{hyperref}
\usepackage{multicol}
\usepackage{cite}  

\title{\LARGE \bf
MSCMHMST: A traffic flow prediction model based on Transformer 
}

\author{$^{1}$Weiyang Geng, $^{2}$Yiming Pan, $^{3}$Zhecong Xing, $^{4}$Dongyu Liu, $^{5}$Rui Liu, and $^{6}$Yuan Zhu*
\thanks{This study was partially supported by the Key Program of Natural Science Foundation of Inner Mongolia, China (Grant No. 2024ZD27), the Commissioned Key Program of Social Science Foundation of Inner Mongolia, China (Grant No. 2024WTZD02), and the Young Scientists Fund of the National Natural Science Foundation of China (Grant No. 61903205). The authors are responsible for the facts and accuracy of the data presented herein, and the contents of this paper do not necessarily reflect the official views or policies of the funding agencies. }
\thanks{$^{1}$Weiyang Geng is with the Department of Computer Science, Inner Mongolia University, Hohhot, Inner Mongolia, 010021, China {\tt\small 32209071@mail.imu.edu.cn}.}%
\thanks{$^{2}$Yiming Pan is with the Department of Transportation Engineering, Inner Mongolia University, Hohhot, Inner Mongolia, 010020, China {\tt\small 0203123327@mail.imu.edu.cn}.}%
\thanks{$^{3}$Zhecong Xing is with the Department of Computer Science, Inner Mongolia University, Hohhot, Inner Mongolia, 010021, China {\tt\small 32209072@mail.imu.edu.cn}.}%
\thanks{$^{4}$Dongyu Liu is with the Department of Computer Science, Inner Mongolia University, Hohhot, Inner Mongolia, 010021, China {\tt\small ldyaky123@163.com}.}%
\thanks{$^{5}$Rui Liu is with the Department of Computer Science, Inner Mongolia University, Hohhot, Inner Mongolia, 010021, China {\tt\small 32209229@mail.imu.edu.cn}.}%
\thanks{$^{6}$Yuan Zhu, Ph.D., is with the Inner Mongolia Center for Transportation Research, Inner Mongolia University, Hohhot, Inner Mongolia, 010020, China {\tt\small zhuyuan@imu.edu.cn} (corresponding author).}%
}

\begin{document}

\maketitle
\thispagestyle{empty}
\pagestyle{empty}

\begin{abstract}

This study proposes a hybrid model based on Transformers, named MSCMHMST, aimed at addressing key challenges in traffic flow prediction. Traditional single-method approaches show limitations in traffic prediction tasks, whereas hybrid methods, by integrating the strengths of different models, can provide more accurate and robust predictions. The MSCMHMST model introduces a multi-head, multi-scale attention mechanism, allowing the model to parallel process different parts of the data and learn its intrinsic representations from multiple perspectives, thereby enhancing the model's ability to handle complex situations. This mechanism enables the model to capture features at various scales effectively, understanding both short-term changes and long-term trends. Verified through experiments on the PeMS04/08 dataset with specific experimental settings, the MSCMHMST model demonstrated excellent robustness and accuracy in long, medium, and short-term traffic flow predictions. The results indicate that this model has significant potential, offering a new and effective solution for the field of traffic flow prediction. 

\end{abstract}

\section{Introduction}

As the core task of intelligent transportation system, traffic flow prediction accuracy directly affects the optimization efficiency of traffic control strategy. Traditional predictive paradigms such as SARIMA[1], ARIMA[2] and SVR[3] have significant theoretical limitations in describing complex spatiotemporal nonlinear correlations. STGNN[4], STGCN[5], Graph WaveNet[6] and other models based on deep learning construct spatial dependency through graph convolutional networks. The CNN-GRU[7], CNN-LSTM[8] and BiLSTM[9] hybrid architectures attempt to combine the spatial perception of convolutional networks with the temporal memory advantages of cyclic networks. However, the existing methods still face two technical bottlenecks: First, the fixed-scale convolution kernel is difficult to adapt to the feature extraction requirements of multi-level traffic modes; Second, there are dual constraints of efficiency and accuracy in long range time-dependent modeling.

In recent years, hybrid architecture innovation has provided a new path for performance breakthrough: Transformer[10] architecture has gained wide attention in the field of traffic prediction due to its advantages in sequence modeling. For example, PDFormer[11] enhances dynamic dependency modeling through transmission delay sensing mechanism, but there is still room for optimization in the multi-scale feature fusion dimension. STAEformer[12] adopted adaptive embedding to improve the prediction accuracy, but its computational complexity increased with the square of the sequence length, which restricted the engineering practicability. It is noteworthy that ST-Transformer[13] uses time-space separation attention step modeling strategy to reduce computing load, but it has insufficient adaptability to dynamic spatial relationships (such as temporary traffic control). TimeSformer[14] adopted spatio-temporal block global self-attention to realize joint modeling, but ignored the inherent spatial locality prior of traffic data. T-Transformer[15] uses graph structure coding to enhance spatial topology perception, but its static graph hypothesis is difficult to adapt to the dynamic evolution of the road network.

The current research system has common technical defects: at the dimension level of spatial modeling, fixed-size convolution kernel is difficult to cooperatively capture microscopic road-level congestion characteristics and macroscopic road-level diffusion patterns; At the time modeling dimension, the global attention mechanism has the problems of short-term emergency response lag and computational redundancy. In terms of computational efficiency, the high memory consumption of multi-head attention mechanisms restricts real-time deployment.

In response to the above challenges, this paper proposes the MSCMHMST model, whose innovation is reflected in three dimensional breakthroughs:

Multi-scale convolution module: a hierarchical feature pyramid architecture was constructed, a 3×3 convolution kernel was used to capture microscopic road level state fluctuations, a 5×5/7×7 convolution kernel was used to model mesolevel regional interaction dynamics, and a 9×9 convolution kernel was used to extract macroscopic road network evolution patterns.
Multi-head multi-scale attention mechanism: Design parallel local - global attention head, local head (window <=15) focus on rapid response to emergencies, global head (window >15) track the long-term evolution of periodic rules.
Spatio-temporal decoupling collaborative architecture: CNN-Transformer dual-flow framework is used to realize spatio-temporal feature decoupling, and multi-scale spatio-temporal interaction is realized combined with dynamic feature gating.
MSCMHMST offers three advantages over existing Transformer improvements:

Spatial modeling flexibility: Discard explicit graph structure dependence and extract spatial features through multi-scale convolution adaptive
Multi-granularity feature collaboration: fusion of convolutional kernel diversity (3/5/7/9) and attentional head functional differentiation (local/global) to achieve cross-scale feature fusion
Optimization of computing efficiency: The time-space decoupling strategy is used to reduce the computational complexity of attention and improve the efficiency of long sequence modeling
The core contributions of this paper can be summarized as follows:

1. The first dual-flow traffic prediction architecture combining multi-scale convolution and functionally differentiated attention heads breaks through the traditional single-scale modeling paradigm, extracts local/global spatial features through multi-scale convolution network, and combines multi-head multi-scale Transformer to model burst/cycle time dependence, realizing spatial-temporal feature decoupling and collaborative learning.
2. Dynamic adaptive attention mechanism: Conv-MHA is designed to dynamically generate attention weights by learning conV-MHA, so as to achieve dynamic focusing of spatio-temporal characteristics of different granularity.
3. The superiority of the model was verified on multiple benchmark data sets, especially the medium and long-term prediction indicators reached SOTA.

\section{Problem Definition}

Traffic flow forecasting is a complex task influenced by multiple factors. This paper utilizes historical data to predict future traffic flow.

Let there be a historical data set \(X_H\) containing traffic volume data for each moment in time. The set \(X_H\) can be expressed as:
\begin{equation}
X_H = \{x_1, x_2, \ldots, x_h |\quad x_h \in \mathbb{R}^d \}
\end{equation}

The predicted traffic flow data for a time \(T\) moments in the future is denoted as:
\begin{equation}
X_{H+T} = \{x_{h+1}, x_{h+2}, \ldots, x_{h+t} | \quad t > h \}
\end{equation}

The model is defined as a function \(F(X_H; \theta, \Phi)\) with the historical data set \(X_H\) as the input and the predicted values \(X_{H+T}\) as the output, where \(\theta\) are the parameters obtained through training data, and \(\Phi\) represents hyperparameters (such as learning rate, batch size, iterations, etc.). This can be expressed as:
\begin{equation}
F(X_H; \theta, \Phi) = X_{H+T}
\end{equation}

\begin{figure}[ht]
\centering
\includegraphics[width=\columnwidth]{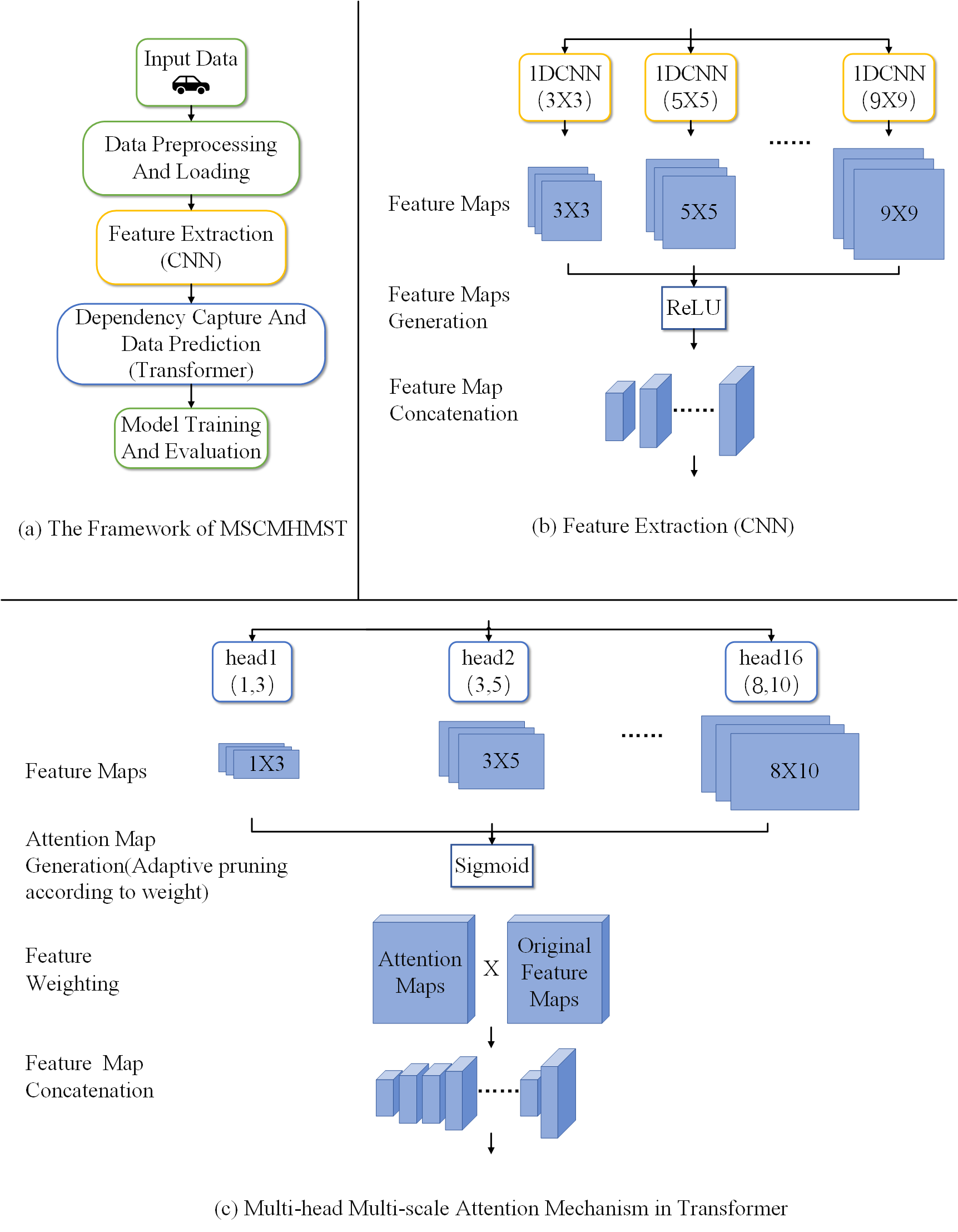}
\caption{Detailed Framework of MSCMHMST. (a) The Framework of MSCMHMST. (b) Feature Extraction (CNN). (c) Multi-head Multi-scale Attention Mechanism in Transformer.}
\end{figure}

Expanding the formula for every moment \(H\), there is a corresponding predicted value for time \(T\) moments later, leading to:

\begin{equation}
\begin{aligned}
F(X_H; \theta, \Phi) &=  \{x_{h+1}, x_{h+2}, \ldots, x_{h+t}\}
\end{aligned}
\end{equation}

Regarding model training and tuning, the primary objective is to find the parameters \(\theta\) that minimize the Loss, which can be denoted as:
\begin{equation}
\theta^* = \arg\min_{\theta} \text{Loss}(F(X_H; \theta, \Phi), X_H)
\end{equation}

Hyperparameter optimization can be conducted through methods such as grid search or Bayesian optimization, represented as:
\begin{equation}
\Phi^* = \arg\min_{\Phi} \text{Loss}(F(X_H; \theta, \Phi), X_H)
\end{equation}

Finally, MAE, MSE, RMSE, and MAPE are chosen as evaluation metrics for the predictive results.

\section{Methodology}

In this section, the specific structure and formula principle of the model proposed in this paper will be introduced in detail.

\subsection{The Framework of MSCMHMST} 
Traffic data often exhibit complex spatiotemporal dependencies. To effectively learn these characteristics and enhance prediction accuracy, the MSCMHMST model aims to achieve the following objectives:

(1)Understand the spatiotemporal dependency features of input traffic data to ensure an accurate grasp of the complex spatiotemporal relationships between data points.

(2)Utilize the overall spatiotemporal correlations of data to improve the predictive precision of future states.

(3)Consider the dependency between short-term fluctuations and long-term trends in traffic flow, focusing on the associations within different time intervals.

The model architecture proposed in the study, as illustrated in Fig. 1(a), initially preprocesses the input data before feeding it into the model for processing. In order to optimize the model's spatiotemporal dependence learning ability, the model is divided into two key parts: convolutional neural networks (CNN) and sequence prediction. The CNN component first extracts spatial features of the traffic flow, which are then fed into the Transformer model. The Transformer model aims to capture temporal dependencies through changing spatial features. Ultimately, the entire model is trained and evaluated to verify its performance. The CNN component is primarily used to extract instantaneous spatial features, which are then input into the Transformer. The Transformer, through the proposed multi-head multi-scale attention mechanism, simultaneously learns both local and global features over time, thus establishing temporal connections between the spatial features. Ultimately, this achieves the spatio-temporal association of traffic flow data.

\subsection{Multiscale Convolution Mechanism}

Multiscale convolution effectively captures multilevel feature information. By utilizing convolutional kernels of various sizes, the network can capture features at different levels of abstraction, aiding the model in recognizing and learning complex patterns within the data. For time series data, this means the model can learn both short-term and long-term dependencies simultaneously. By combining convolutional kernels of different sizes, the network can perceive broader contextual information while maintaining high spatial resolution, which is particularly beneficial for understanding sequential data such as traffic flow. Multiscale convolution improves the model's adaptability to variations in input data by providing information across different scales, enabling the model to better handle various traffic scenarios from local small-scale incidents to overall trend changes. The CNN model's details are referenced in Fig. 1(b). The preprocessed data is input and feature extraction is performed using multiple 1D CNNs of different scales, generating feature maps for each scale which are then concatenated.

The principle of the multiscale CNN model is as follows: Multiscale convolution blocks perform convolution operations on the input feature map with convolutional kernels of different sizes to capture features at various scales. Suppose the input feature map is denoted as \(X \in \mathbb{R}^{C_{\text{in}} + L}\) where \(C_{\text{in}}\)

\noindent is the number of input channels and \(L\) is the feature length (or time steps).

For each convolutional kernel size \(k\), the convolution operation can be expressed as:
\begin{equation}
Y_k^{(i)} = \sum_{m=0}^{k-1} W_k[m] \cdot X_{i+m-\frac{k}{2}} + b_k \quad 
\end{equation}
where:
\begin{itemize}
    \item \(Y_k\) is the output feature map after applying the convolutional kernel \(k\).
    \item \(W_k[m]\) is the weight of the convolutional kernel \(k\) at position \(m\).
    \item \(b_k\) is the bias term associated with the convolutional kernel \(k\).
    \item \(i\) is the position index of the output feature map.
    \item \(\frac{k}{2}\) is the offset to ensure the convolution operation remains centered.
\end{itemize}

Then, for each output feature map \(Y_k\), the study applies the ReLU activation function to increase nonlinear processing capability:
\begin{equation}
A_k^{(i)} = \max(0, Y_k^{(i)}) \quad 
\end{equation}

Finally, the output feature maps from convolutional kernels of different sizes are fused along the channel dimension to form the final output feature map \(Z\) expressed as:
\begin{equation}
Z = \bigoplus_k A_k \quad 
\end{equation}
where \(\bigoplus\) denotes the concatenation operation along the channel dimension.

\subsection{Multi-head Multi-scale Attention Mechanism}

Within the Transformer model, this paper highlights the proposed multi-head multi-scale attention mechanism. This mechanism allows the model to process different parts of the data in parallel and learn the intrinsic representation of data from multiple perspectives. It enhances the model's capability to handle complex scenarios such as considering various traffic factors and environmental variables simultaneously. The multi-scale approach of each head enables the model to capture features at different scales, recognizing and utilizing both local details and global patterns within the data. For instance, in traffic flow prediction, the model can concurrently comprehend short-term changes (like instant congestion) and long-term trends (such as variations during rush hours). The main structure of attention is referred to in Fig. 1(c).

During feature extraction, each head \( \text{head}_i \) utilizes a set of convolutional kernels \( k_j \in K \) to extract feature maps \( F_{ij} \), which is described by equation (10):
\begin{equation}
F_{ij} = \text{Conv}_{k_j}(X) \quad 
\end{equation}
where \( \text{Conv}_{k_j} \) is the convolution operation and \( X \) is the input data.

Next, the spatio-temporal dependence of feature graph \( F_{ij} \) is extracted by \( \text{head}_i \) mechanism, and the attention map \( A_{ij} \) is generated dynamically by multi-scale convolution kernel 
\begin{figure}[ht]
\centering
\includegraphics[width=\columnwidth]{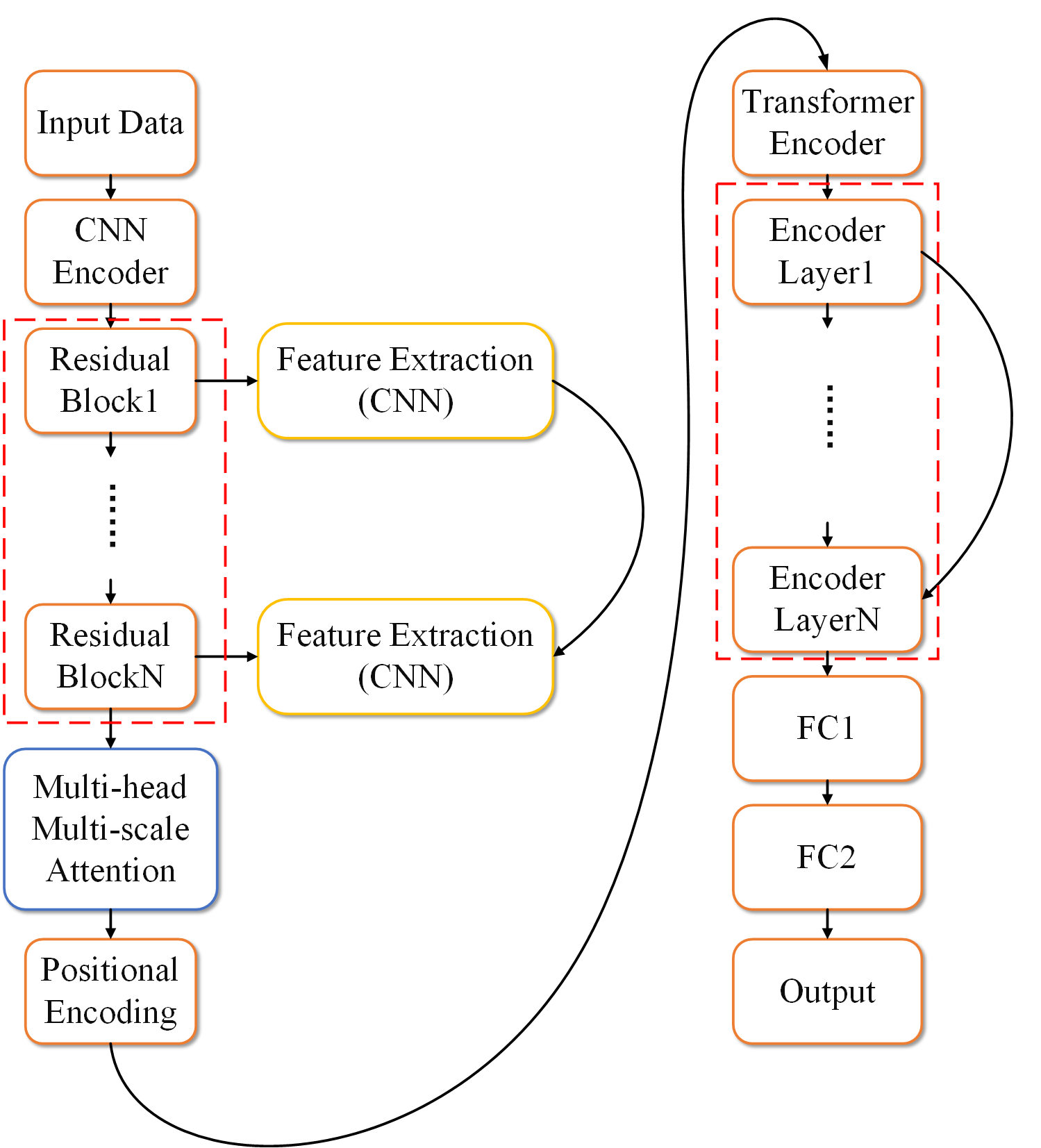}
\caption{Network of MSCMHMST}
\end{figure}

\noindent \( k_j \) , and the feature importance is quantified by activation function \( \sigma \) , The dynamic compression of model redundancy features is realized by adaptive pruning low response region, which takes into account multi-scale feature preservation and computational efficiency optimization. expressed as equation (11):
\begin{equation}
A_{ij} = \sigma(\text{Conv}_{k_j}^{\text{head}_i}(F_{ij})) \quad 
\end{equation}
where \( \sigma \) is the Sigmoid activation function, and \( \text{Conv}_{k_j}^{\text{head}_i} \) represents the convolution operation corresponding to head \( \text{head}_i \) and scale \( k_j \). Subsequently, for feature weight allocation, each feature map \( F_{ij} \) is weighted by its corresponding attention map \( A_{ij} \) denoted as (12):
\begin{equation}
W_{ij} = A_{ij} \odot F_{ij} \quad 
\end{equation}
where \( \odot \) represents the Hadamard product (element-wise multiplication).\par

The feature maps are then concatenated. For each head \( \text{head}_i \), all weighted feature maps \( W_{ij} \) are concatenated along the channel dimension as shown in equation (13):
\begin{equation}
H_i = \bigoplus_{j \in K} W_{ij} \quad 
\end{equation}

Finally, the outputs of all heads \( H_i \) are concatenated along the channel dimension to form the final multi-head multi-scale feature representation \( \text{MHMS} \) as illustrated in (14):
\begin{equation}
MHMS = \bigoplus_{i \in \text{Heads}} H_i \quad 
\end{equation}
where \( \text{Heads} \) denotes all the heads.

\subsection{Network Structure of MSCMHMST}
This section introduces the proposed hybrid CNN-Transformer model architecture, designed to process complex time series data effectively by capturing both local features and global dependencies. The model combines the strengths of CNN and Transformer technologies to enhance the accuracy and efficiency of time series predictions.

Fig .2, The model starts with a Multi-Scale Convolutional Encoder (CNNEncoder), featuring a Multi-Scale Convolution Block (MultiScaleConvBlock) for capturing multi-level features through convolutional kernels of varying sizes, enhanced by optional residual connections to prevent gradient vanishing and ensure deep network information flow.

Following, the Multi-Head Multi-Scale Attention mechanism further processes features encoded by the CNNEncoder, distributing data across multiple "heads" with different scale convolutional kernels for comprehensive input analysis. This enables focused attention on crucial data parts, boosting processing efficiency and predictive accuracy.

Positional Encoding is introduced to allow the Transformer part to understand the temporal order within sequence data, a crucial aspect for maintaining the time characteristics of the data.

The model then utilizes a Transformer Encoder (TransformerEncoder) with multiple layers comprising self-attention mechanisms, feed-forward neural networks, residual connections, and layer normalization to capture long-distance dependencies between input features, enhancing global information processing.

Finally, a fully connected layer (Linear FC1) and an output layer (Output Linear FC2) map the high-dimensional features to prediction targets, converting the complex Transformer encoder representations to specific predictive values, thus completing the process from data input to prediction output.

\section{Experiment}

\subsection{Dataset}

To validate the performance of the MSCMHMST model, the study conducted extensive experiments on the PeMS04 and PeMS08 datasets, which are based on real-world traffic data collection. These datasets are collected by the Performance Measurement System (PeMS) from the California Department of Transportation. The traffic data is aggregated at 5-minute intervals, resulting in 288 data points per day [16]. 

\subsection{Comparison Model}

A total of 21 models were compared in this paper, including 12 baseline models (ARIMA, SARIMA, SVR, LSTM, GRU, 1DCNN\_LSTM, STGCN, Transformer, TimeSformer, ST-Transformer, T-Transformer). To evaluate the performance of the model. 9 ablation comparison models (MSCMHMST\_4, MSCMHMST\_8, MSCMHMST\_16, 1DCNN\_Transformer, 1DCNN\_MHMST, MSC\_Transform-er, MSC1R\_MHMST1L, MSC2R\_MHMST2L, MSC3R\_M-HMST3L). Among them, MSCMHMST\_4/8/16, based on multi-head multi-scale attention mechanism (MHMST), is configured with 4/8/16 attention heads respectively. 1DCNN\_Transformer, 1D convolution layer + standard Transformer encoder; 1DCNN\_MHMST, one-dimensional convolutional layer +MHMST module; MSC\_Transformer, Multi-scale Convolution (MSC) module + standard Transformer; MSC1R\_MHMST1L, introduces a residual block and an additional single-layer decoder layer (mainly composed of residual blocks). The MSC2R\_MHMST2L and MSC3R\_MHMST3L models further increase the number of layers on this basis.

\subsection{Evaluation}
The performance of each model was evaluated by four indicators (MAE, MSE, RMSE, MAPE). The formula is as follows:
\[
\text{MAE} = \frac{1}{n} \sum_{i=1}^{n} \left| y_i - \hat{y}_i \right|
\]

\begin{table*}[!htbp]
\footnotesize 
\setlength{\tabcolsep}{5pt} \centering
\caption{Performance comparison of different models with 3, 6, and 12 steps on datasets PeMS04 and PeMS08.}
\resizebox{\textwidth}{!}{%
    \begin{tabular}{lllllllllllllllllll}
    \hline
        Dataset & Type & ~ & Model & ~ & 15m(3step) & ~ & ~ & ~ & ~ & 30m(6step) & ~ & ~ & ~ & ~ & 60m(12step) & ~ & ~ & ~ \\ \hline
        ~ & ~ & ~ & ~ & ~ & MAE & MSE & RMSE & MAPE & ~ & MAE & MSE & RMSE & MAPE & ~ & MAE & MSE & RMSE & MAPE \\ \hline
        PeMS04 & Statistical & ~ & ARIMA & ~ & 51.21  & 5551.50  & 74.51  & 26.99  & ~ & 46.78  & 4878.88  & 69.85  & 24.22  & ~ & 49.06  & 5199.33  & 72.11  & 24.27  \\ 
        ~ & ~ & ~ & SARIMA & ~ & 49.63  & 5408.68  & 73.54  & \bf{23.51}  & ~ & 49.63  & 5409.02  & 73.55  & 23.52  & ~ & 49.64  & 5409.33  & 73.55  & 23.52  \\ 
        ~ & Machine Learning & ~ & SVR & ~ & 52.88  & 5520.05  & 74.30  & 30.32  & ~ & 53.31  & 5585.99  & 74.74  & 30.95  & ~ & 56.96  & 6113.26  & 78.19  & 35.59  \\ 
        ~ & Deep Learning & ~ & 1DCNN\_LSTM & ~ & 46.93  & \bf{4463.90}  & \bf{66.81}  & 23.95  & ~ & 46.07  & 4479.10  & 66.93  & 23.80  & ~ & 46.93  & 4641.07  & 68.12  & 23.91  \\ 
        ~ & ~ & ~ & STGNN & ~ & 47.02  & 4523.88  & 67.26  & 24.48  & ~ & 48.03  & 4643.10  & 68.13  & 26.59  & ~ & 49.64  & 4838.67  & 69.56  & 25.27  \\ 
        ~ & ~ & ~ & STGCN & ~ & 49.21  & 4685.89  & 68.45  & 25.26  & ~ & 49.41  & 4683.09  & 68.43  & 25.33  & ~ & 48.16  & 4606.50  & 67.87  & 24.80  \\ 
        ~ & ~ & ~ & LSTM & ~ & 48.07  & 4602.28  & 67.84  & 24.42  & ~ & 47.51  & 4564.49  & 67.56  & 24.27  & ~ & 47.10  & 4478.74  & 66.92  & 23.85  \\ 
        ~ & ~ & ~ & GRU & ~ & 47.37  & 4640.99  & 68.12  & 25.17  & ~ & 47.88  & 4544.27  & 67.41  & 24.18  & ~ & 46.92  & 4496.80  & 67.06  & 23.99  \\ 
        ~ & ~ & ~ & Transformer & ~ & 55.41  & 6068.93  & 77.82  & 36.88  & ~ & 55.57  & 6127.23  & 78.19  & 39.97  & ~ & 55.80  & 6253.89  & 79.00  & 40.66  \\ 
        ~ & ~ & ~ & TimeSformer & ~ & 49.43  & 4624.81  & 68.00  & 24.98  & ~ & 48.91  & 4488.61  & 67.00  & 24.42  & ~ & 47.32  & 4497.89  & 67.06  & 23.95  \\ 
        ~ & ~ & ~ & ST-Transformer & ~ & 49.53  & 4641.82  & 68.13  & 25.09  & ~ & 49.02  & 4513.60  & 67.18  & 24.79  & ~ & 47.98  & 4534.80  & 67.34  & 24.23  \\ 
        ~ & ~ & ~ & T-Transformer & ~ & 50.41  & 4671.71  & 68.35  & 25.47  & ~ & 48.39  & 4468.77  & 66.85  & 24.30  & ~ & 48.75  & 4534.89  & 67.34  & 24.34  \\ 
        ~ & ~ & ~ & \bf{MSCMHMST} & ~ & \bf{46.76}  & 4492.01  & 67.02  & 24.19  & ~ & \bf{45.27}  & \bf{4325.24}  & \bf{65.77}  & \bf{23.26}  & ~ & \bf{45.31}  & \bf{4265.96}  & \bf{65.31}  & \bf{22.84}  \\ \hline
        PeMS04 & Ablation Experimental Model & ~ & MSCMHMST\_4 & ~ & 47.02  & 4534.94  & 67.34  & 24.27  & ~ & \bf{45.15}  & 4330.28  & 65.80  & 23.42  & ~ & 45.14  & 4291.38  & 65.50  & 23.45  \\ 
        ~ & ~ & ~ & MSCMHMST\_8 & ~ & 46.75  & 4518.70  & 67.22  & 24.26  & ~ & 46.00  & 4406.06  & 66.38  & 23.63  & ~ & \bf{45.04}  & \bf{4241.02}  & \bf{65.12}  & \bf{22.81}  \\ 
        ~ & ~ & ~ & MSCMHMST\_16 & ~ & 46.76  & \bf{4492.01}  & \bf{67.02}  & 24.19  & ~ & 45.27  & \bf{4325.24}  & \bf{65.77}  & \bf{23.26}  & ~ & 45.31  & 4265.96  & 65.31  & 22.84  \\ 
        ~ & ~ & ~ & 1DCNN\_MHMST & ~ & 46.61  & 4536.56  & 67.35  & 24.57  & ~ & 46.45  & 4446.36  & 66.68  & 23.73  & ~ & 45.97  & 4339.43  & 65.87  & 23.63  \\ 
        ~ & ~ & ~ & 1DCNN\_Transformer & ~ & 46.77  & 4500.63  & 67.08  & \bf{23.97}  & ~ & 46.37  & 4448.62  & 66.70  & 23.81  & ~ & 45.94  & 4439.13  & 66.62  & 23.83  \\ 
        ~ & ~ & ~ & MSC\_Transformer & ~ & \bf{46.40}  & 4587.52  & 67.73  & 25.39  & ~ & 46.94  & 4569.19  & 67.59  & 24.95  & ~ & 47.12  & 4615.96  & 67.94  & 25.52  \\ 
        ~ & ~ & ~ & MSC1R\_MHMST1L & ~ & 47.13  & 4511.39  & 67.17  & 24.32  & ~ & 46.05  & 4488.74  & 67.00  & 23.96  & ~ & 48.90  & 5124.89  & 71.55  & 26.28  \\ 
        ~ & ~ & ~ & MSC2R\_MHMST2L & ~ & 46.80  & 4573.66  & 67.63  & 24.41  & ~ & 46.72  & 4592.20  & 67.76  & 23.79  & ~ & 50.11  & 5318.64  & 72.91  & 25.14  \\ 
        ~ & ~ & ~ & MSC3R\_MHMST3L & ~ & 46.68  & 4528.84  & 67.30  & 24.19  & ~ & 46.60  & 4550.08  & 67.45  & 23.84  & ~ & 50.03  & 5279.41  & 72.62  & 25.45  \\ \hline
        PeMS08 & Statistical & ~ & ARIMA & ~ & 22.24  & 893.96  & 29.90  & 19.36  & ~ & 20.04  & 758.72  & 27.54  & 17.28  & ~ & 20.87  & 819.30  & 28.62  & 18.05  \\ 
        ~ & ~ & ~ & SARIMA & ~ & 22.47  & 938.33  & 30.63  & 18.83  & ~ & 22.48  & 938.53  & 30.64  & 18.83  & ~ & 22.46  & 938.14  & 30.63  & 18.80  \\ 
        ~ & Machine Learning & ~ & SVR & ~ & 23.72  & 994.55  & 31.54  & 21.08  & ~ & 23.41  & 970.39  & 31.15  & 21.14  & ~ & 25.28  & 1108.63  & 33.30  & 23.91  \\ 
        ~ & Deep Learning & ~ & 1DCNN\_LSTM & ~ & \bf{20.26}  & \bf{773.46}  & \bf{27.81}  & 14.61  & ~ & 19.91  & 750.60  & 27.40  & 14.32  & ~ & 20.00  & 749.72  & 27.38  & 14.41  \\ 
        ~ & ~ & ~ & STGNN & ~ & 20.60  & 786.36  & 28.04  & 14.83  & ~ & 20.02  & 763.60  & 27.63  & 14.72  & ~ & 21.88  & 871.95  & 29.53  & 16.45  \\ 
        ~ & ~ & ~ & STGCN & ~ & 20.72  & 793.91  & 28.18  & 15.06  & ~ & 20.35  & 768.71  & 27.72  & 14.71  & ~ & 20.06  & 749.51  & 27.38  & 14.54  \\ 
        ~ & ~ & ~ & LSTM & ~ & 20.61  & 804.90  & 28.37  & 14.91  & ~ & 20.30  & 780.02  & 27.93  & 14.69  & ~ & 20.08  & 761.71  & 27.60  & 14.54  \\ 
        ~ & ~ & ~ & GRU & ~ & 20.66  & 803.37  & 28.34  & 14.91  & ~ & 20.23  & 776.40  & 27.86  & 14.71  & ~ & 20.03  & 756.28  & 27.50  & 14.56  \\ 
        ~ & ~ & ~ & Transformer & ~ & 25.89  & 1226.11  & 35.00  & 20.92  & ~ & 27.12  & 1361.18  & 36.85  & 21.28  & ~ & 26.06  & 1264.38  & 35.51  & 21.06  \\ 
        ~ & ~ & ~ & TimeSformer & ~ & 20.73  & 791.04  & 28.13  & 15.06  & ~ & 20.34  & 755.05  & 27.48  & 14.79  & ~ & 20.25  & 758.07  & 27.53  & 14.50  \\ 
        ~ & ~ & ~ & ST-Transformer & ~ & 21.00  & 799.52  & 28.28  & 15.24  & ~ & 20.60  & 765.92  & 27.67  & 14.80  & ~ & 20.21  & 742.34  & 27.25  & 14.62  \\ 
        ~ & ~ & ~ & T-Transformer & ~ & 20.71  & 794.58  & 28.19  & 15.07  & ~ & 20.32  & 755.64  & 27.49  & 14.73  & ~ & 20.15  & 744.57  & 27.28  & 14.54  \\ 
        ~ & ~ & ~ & \bf{MSCMHMST} & ~ & 20.37  & 789.46  & 28.10  & \bf{14.60}  & ~ & \bf{19.77}  & \bf{743.89}  & \bf{27.27}  & \bf{14.24}  & ~ & \bf{19.74}  & \bf{730.67}  & \bf{27.03}  & \bf{14.30}  \\ \hline
        PeMS08 & Ablation Experimental Model & ~ & MSCMHMST\_4 & ~ & 20.33  & 788.85  & 28.09  & 14.64  & ~ & 19.84  & 750.53  & 27.40  & 14.32  & ~ & 19.72  & 737.49  & 27.16  & 14.27  \\ 
        ~ & ~ & ~ & MSCMHMST\_8 & ~ & \bf{20.23}  & 782.53  & 27.97  & 14.62  & ~ & 19.82  & 749.59  & 27.38  & 14.37  & ~ & \bf{19.68}  & 735.37  & 27.12  & \bf{14.22}  \\ 
        ~ & ~ & ~ & MSCMHMST\_16 & ~ & 20.37  & 789.46  & 28.10  & \bf{14.60}  & ~ & \bf{19.77}  & \bf{743.89}  & \bf{27.27}  & \bf{14.24}  & ~ & 19.74  & \bf{730.67}  & \bf{27.03}  & 14.30  \\ 
        ~ & ~ & ~ & 1DCNN\_MHMST & ~ & 20.51  & 799.92  & 28.28  & 14.87  & ~ & 20.08  & 764.76  & 27.65  & 14.63  & ~ & 19.74  & 734.62  & 27.10  & 14.30  \\ 
        ~ & ~ & ~ & 1DCNN\_Transformer & ~ & 20.54  & 791.78  & 28.14  & 14.77  & ~ & 19.91  & 745.49  & 27.30  & 14.58  & ~ & 19.94  & 744.23  & 27.28  & 14.55  \\ 
        ~ & ~ & ~ & MSC\_Transformer & ~ & 21.11  & 838.85  & 28.96  & 15.45  & ~ & 20.81  & 812.25  & 28.50  & 15.04  & ~ & 21.05  & 823.96  & 28.70  & 15.46  \\ 
        ~ & ~ & ~ & MSC1R\_MHMST1L & ~ & 20.48  & 793.30  & 28.17  & 14.75  & ~ & 20.10  & 766.15  & 27.68  & 14.70  & ~ & 20.96  & 819.36  & 28.62  & 15.52  \\ 
        ~ & ~ & ~ & MSC2R\_MHMST2L & ~ & 20.43  & 788.13  & 28.07  & 14.68  & ~ & 20.17  & 765.27  & 27.66  & 14.54  & ~ & 22.89  & 965.41  & 31.06  & 17.15  \\ 
        ~ & ~ & ~ & MSC3R\_MHMST3L & ~ & 20.28  & \bf{782.17}  & \bf{27.97}  & 14.70  & ~ & 20.41  & 780.98  & 27.95  & 14.66  & ~ & 22.27  & 927.90  & 30.45  & 16.38 \\ \hline
    \end{tabular}
}
\end{table*}

\[
\text{MSE} = \frac{1}{n} \sum_{i=1}^{n} \left( y_i - \hat{y}_i \right)^2
\]
\[
\text{RMSE} = \sqrt{\frac{1}{n} \sum_{i=1}^{n} \left( y_i - \hat{y}_i \right)^2}
\]
\[
\text{MAPE} = \frac{100\%}{n} \sum_{i=1}^{n} \left| \frac{y_i - \hat{y}_i}{y_i} \right|
\]
where:
\begin{itemize}
    \item \(y_i\) is Actual values.
    \item \(\hat{y}_i\) is Predicted values.
    \item \(n\) is Number of observations.
\end{itemize}

\subsection{Experimental Setup}

The optimal hyperparameters are determined by multiple tracking error tests. The hyperparameters are set as follows: the batch size is set to 32, the learning rate is set to 0.001, the number of training epochs is set to 100, the Adam optimizer [17] is used, the number of hidden layer elements is set to 8, the number of multi-scale convolution kernel is set to 4, and the size of multi-scale convolution kernel is set to 3, 5, 7, 9. Note that the number of heads is set to 16, 1 to 16 head [1, 3], [3, 5], [5, 7], [7, 9], [1, 5], [3, 7], [5, 9], [1, 7], [1, 9], [2, 6], [4, 8], [3, 9], [2, 4], [4, 6], [6, 8], [8, 10] of the scale. The experiment was conducted on NVIDIA Quadro RTX 3000 graphics card with PyTorch 2.3.1 and Python 3.8.19 to ensure the repeatability of the experiment.

In the training set, PeMS04 contains data for 44 days (a total of 12,672 time steps), while PeMS08 contains data for 47 days (a total of 13,536 time steps). The validation and test sets are the same for both datasets, with the validation set covering 5 days (a total of 1,440 time steps) and the test set covering 10 days (a total of 2,880 time steps).

To ensure fairness in the experiments, this study tested various configurations with hidden layers of 8, 16, 32, 64, and 128. All models were set with the optimal average results to explore performance differences among them. Some parameters of the neural network models could be reused, while the statistical models were selected with the optimal parameters.

\subsection{Experimental Result}

In order to ensure the statistical stability of the experimental conclusions, repeated experiments and outlier elimination strategies were adopted in this study: 10 independent repeated experiments were conducted on the PeMS04/PeMS08 dataset, and the 2 experiments with the largest deviation in each task (Z-score>2.5) were eliminated, and the robustness evaluation was conducted based on the mean of the remaining 8 results. The performance of MSCMHMST on the PeMS04/08 dataset is as follows:

(1) the medium and long term prediction (30-60 minutes) has a significant advantage, and the 6/12-step prediction index is ahead, while the 3-step prediction has a relatively weak advantage due to the limited short-term temporal context; 

(2) On the PeMS04 dataset with more complex traffic patterns, the MAE of this model is reduced by about 11\% (from 56.96 to 45.31) compared with the baseline model SVR, which verifies its adaptability and generalization ability to complex scenarios.

\subsection{Ablation Study}
Additionally, to explore the potential applications of Transformer models in traffic flow prediction, a series of experiments were conducted to evaluate model performance. These experiments aimed to assess whether hybrid models perform better than single models, whether multi-scale convolution can enhance model performance, the effectiveness of multi-head multi-scale attention mechanisms, and whether the MSCMHMST model can improve overall performance by increasing design complexity. Based on the experimental results as follows:

1. Influence of the number of multi-attention heads: MSCMHMST\_16 (16 heads) performed best in the 60-minute prediction of PEMS04/08 (e.g. RMSE=65.31 on PeMS04, which was 0.4-0.8\% lower than that of 4-8 heads), indicating that multi-head mechanism can effectively extract multi-scale spatio-temporal features. It is worth noting that, thanks to the adaptive head pruning mechanism, MSCMHMST\_4 dynamically allocates computing resources during training, requiring only about 50\% of the training time of the 16-head and 8-head model (The 16-head and 8-head models take the same amount of time), and its performance is very close to that of MSCMHMST\_8 (difference <0.5\%) and MSCMHMST\_16 (difference <1.2\%). This phenomenon verifies the effectiveness of pruning mechanism in balancing model efficiency and accuracy, which can significantly reduce the calculation cost while guaranteeing the prediction performance.

2. Module validity:
MSC module: MSC\_Transformer (without MHMST) has significantly decreased performance in the long term prediction (MAE=47.12 for 60 minutes on PeMS04 vs. 45.31 for MSCMHMST\_16), verifying the necessity of multi-scale convolution and MHMST collaboration.
MHMST substitution: 1DCNN\_MHMST (without MSC) had a 60-minute MAE=45.97 on PeMS04, which was weaker than MSCMHMST\_16 (45.31), demonstrating the advantage of multi-scale convolution in timing modeling.

3. Influence of structural depth: Increasing the number of residual blocks and decoding layers (MSC1R/2R/3R series) leads to a decline in long-term prediction performance (for example, MSC3R's 60-minute MAE=50.03 on PeMS04, which is 10.4\% higher than MSCMHMST\_16↑), indicating that too deep structure may lead to optimization difficulties.

\section{Conclusion}

This study proposes a novel traffic flow prediction model that integrates multi-head multi-scale attention mechanisms with convolutional methods. This combination effectively addresses both local and global variations in traffic data, ensuring high prediction accuracy across different time periods. Test results demonstrate that the model performs exceptionally well on the PeMS04 and PeMS08 datasets. Compared to traditional and deep learning models, it exhibits outstanding performance and high adaptability, particularly in medium to long-term (30 to 60 minutes) traffic flow prediction scenarios. Ablation studies reveal the potential of Transformer applications in this context and emphasize the need for a careful balance between model complexity and performance. Despite the model’s impressive performance, further optimization is required based on specific data types and network sizes. This research provides a powerful predictive tool for intelligent transportation systems and lays a foundation for future research.

\addtolength{\textheight}{-12cm}   





\end{document}